\documentclass[letterpaper]{article} % DO NOT CHANGE THIS
\usepackage[preprint]{aaai2027}
\usepackage[hyphens]{url}  % DO NOT CHANGE THIS
\usepackage{graphicx} % DO NOT CHANGE THIS
\usepackage{natbib}  % DO NOT CHANGE THIS AND DO NOT ADD ANY OPTIONS TO IT
\usepackage{caption} % DO NOT CHANGE THIS AND DO NOT ADD ANY OPTIONS TO IT
\usepackage{url}
\usepackage{xcolor}
\definecolor{venuegray}{gray}{0.55}
\newcommand{\venue}[1]{%
    \hspace{0.15em}{\color{venuegray}\tiny[#1]}%
}
\usepackage{algorithm}
\usepackage{algorithmic}
\usepackage{amsmath}
\usepackage{amssymb}
\usepackage{booktabs}
\usepackage{multirow}

\usepackage{newfloat}
\usepackage{listings}
\DeclareCaptionStyle{ruled}{labelfont=normalfont,labelsep=colon,strut=off} % DO NOT CHANGE THIS
\floatstyle{ruled}
\newfloat{listing}{tb}{lst}{}
\floatname{listing}{Listing}

\usepackage{booktabs}

\title{Core-KAN: Continuous Vision Kernels with Kolmogorov-Arnold Networks}
\author{
    Lan Guo\textsuperscript{\rm 1},
    Mengling Li\textsuperscript{\rm 1},
    Haoran Li\textsuperscript{\rm 2},
    Jun Shen\textsuperscript{\rm 3},
    Yuanbo Jiang\textsuperscript{\rm 1},
    Qingguo Zhou\textsuperscript{\rm 1},
    Binbin Yong\textsuperscript{\rm 1}\thanks{Corresponding author:
    Binbin Yong(yongbb@lzu.edu.cn)}
}

\affiliations{
    \textsuperscript{\rm 1} School of Information Science and Engineering, Lanzhou University\\
    \textsuperscript{\rm 2} Department of Data Science and AI, Monash University\\
    \textsuperscript{\rm 3} School of Computing and Information Technology, University of Wollongong\\
}

\begin{document}

\maketitle

\begin{abstract}
Conventional convolutional kernels are typically defined on fixed discrete grids, limiting their ability to accommodate heterogeneous local structures. Existing adaptive operators improve flexibility, but often couple geometric scale variation with content-dependent filtering, while incurring high computational cost from per-location kernel generation. To decouple geometric scale adaptation from content-dependent filtering while avoiding expensive per-location kernel generation, we propose Continuous Relative-scale KAN (Core-KAN), a relative-scale-conditioned continuous convolution operator. In detail,  Core-KAN maps input features into a compact latent basis space and uses a lightweight scale controller to predict local scales relative to an exponential moving average reference. A KAN-based generator represents depth-wise kernel bases as continuous coordinate functions, allowing the operator to synthesize spatial filters at arbitrary resolutions rather than being confined to a fixed lattice. Instead of synthesizing independent kernels at every location, it constructs a compact bank of scale-conditioned kernel responses and interpolates them according to the predicted local scale map. An independent mixing controller further combines the interpolated basis responses based on local content, explicitly decoupling geometric scale adaptation from content-dependent filtering. Together with lightweight pointwise projections, this design forms a low-rank dynamic convolution that scales efficiently with kernel size and can be readily integrated into hierarchical vision backbones. Extensive experiments across three representative computer vision tasks demonstrate that Core-KAN consistently outperforms strong convolutional and dynamic-kernel baselines while introducing only marginal parameter and computational overhead. Core-KAN provides an efficient and general framework for continuous, scale-adaptive convolution across diverse vision tasks.
\end{abstract}

% \vspace{1em} % 可选：如果需要和上方的 Abstract 拉开一点距离，可以调整这个数值
% \noindent \textbf{Code} --- \url{https://github.com/liml12138/CKKAN}

% Uncomment the following to link to your code, datasets, an extended version or similar.
% You must keep this block between (not within) the abstract and the main body of the paper.
% Make sure that you do not de-anonymize yourself with these links.

\section{Introduction}
\label{sec:introduction}

Local aggregation is a fundamental mechanism for building visual representations. Convolution remains one of the most reliable choices because a small spatially shared kernel provides translation equivariance, computational efficiency, and a strong locality bias \cite{krizhevsky2012alexnet,simonyan2015vgg,he2016deep,liu2022convnext,woo2023convnextv2}. Yet the same property that makes convolution efficient also limits its adaptivity: a fixed discrete kernel profile is applied to textures, object boundaries, thin structures, and homogeneous regions alike. Modern vision backbones therefore face a recurring tension between the efficiency of shared convolutional operators and the need for location-specific spatial reasoning.

\begin{figure}[!t]
    \centering
    \includegraphics[width=0.95\columnwidth]{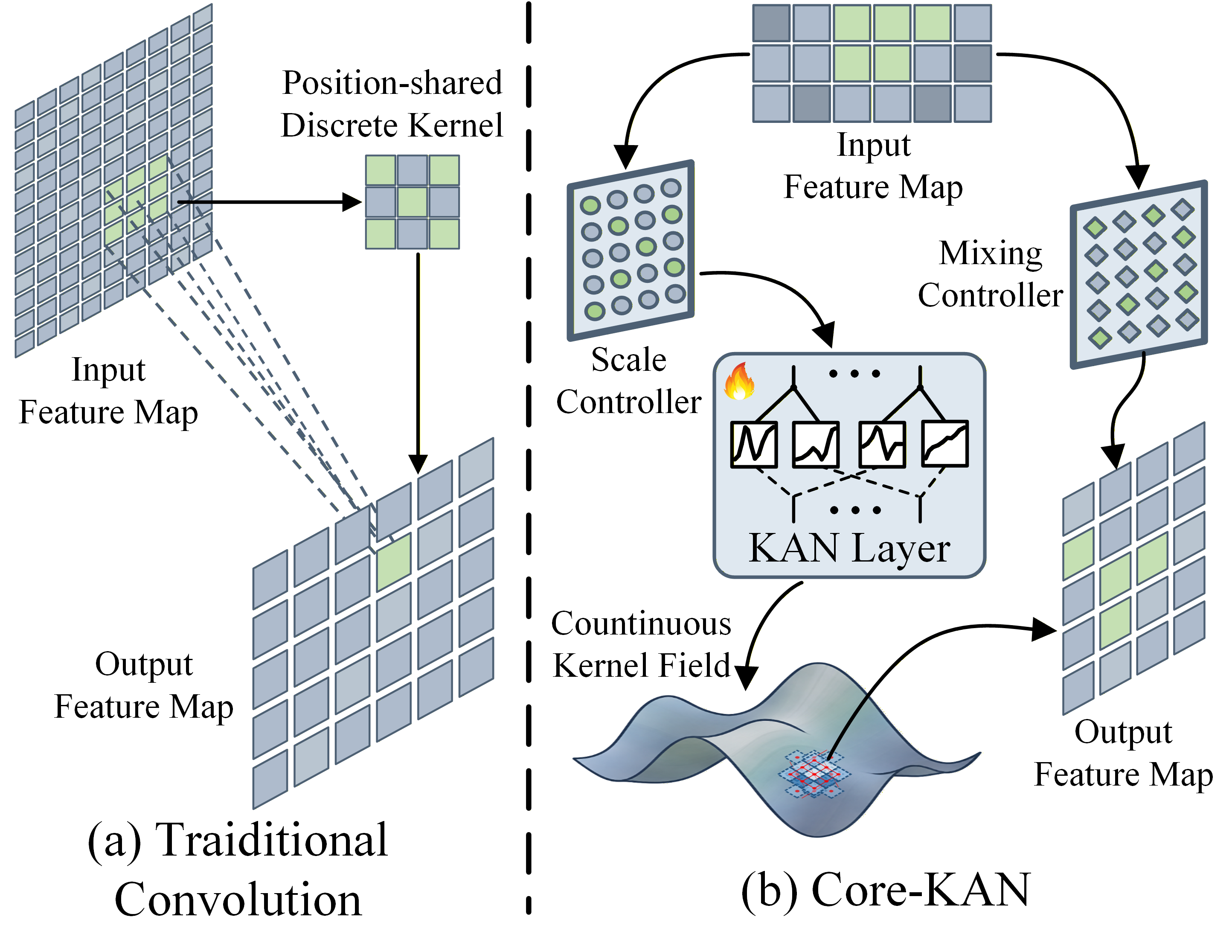}
    \caption{Conceptual comparison between conventional convolution and Core-KAN. (a) Conventional convolution shares a fixed discrete kernel across locations. (b) Core-KAN conditions a shared KAN-parameterized continuous kernel field on local relative scales and independently mixes basis responses according to content, enabling location-specific aggregation on a fixed sampling grid. Implementation details are omitted for clarity.}
    \label{fig:corekan-comparison}
\end{figure}

A direct way to increase adaptivity is to generate or select a different kernel at each location. However, fully position-specific kernels weaken the parameter sharing that makes convolution transferable, and dense kernel synthesis is expensive in both computation and memory. Existing approaches address parts of this tension. Large-kernel and multi-branch networks enlarge or diversify receptive fields, but still operate over finitely parameterized filters \cite{szegedy2015inception,li2019selectivekernel,ding2022replknet,liu2023slak,ding2024unireplknet}. Conditional and dynamic convolutions improve content adaptivity by routing among learned components or modulating kernel weights \cite{yang2019condconv,chen2020dynamicconv,ma2020weightnet,li2022odconv,jia2016dynamicfilter,zhou2021ddf}. Deformable convolutions adapt sampling locations, but still attach weights to a finite set of displaced points \cite{dai2017deformable,zhu2019dcnv2,wang2023internimage,xiong2024dcnv4}. Continuous convolutional representations learn coordinate-to-weight functions, but their kernel fields are typically shared at the layer level rather than conditioned densely on local structure \cite{wang2018pccn,wu2019pointconv,romero2022ckconv,romero2022flexconv}. Thus, an efficient operator that provides dense spatial adaptation through a shared continuous kernel family remains underexplored.

Our key observation is that spatial adaptation in convolution involves two distinct decisions. The first is geometric: local scale and structure may require a sharper, broader, or differently shaped kernel profile. The second is content-dependent: local visual evidence may require a different mixture of latent filtering patterns. Existing dynamic-kernel mechanisms often bind these roles into a single routing signal, reducing interpretability by obscuring whether the operator changes kernel geometry, feature-dependent composition, or merely selects among discrete alternatives. Moreover, absolute scale predictions can drift across feature stages and optimization dynamics, making dense scale conditioning unstable without calibration to a feature-level reference.

Based on this observation, we propose  Continuous Relative-scale KAN (Core-KAN), a Continuous Relative-scale KAN convolution operator for spatially adaptive visual aggregation. As illustrated in Figure~\ref{fig:corekan-comparison}, unlike conventional convolution with a fixed discrete kernel, Core-KAN represents depthwise kernel bases as a shared continuous coordinate field parameterized by a Kolmogorov-Arnold Network (KAN) \cite{liu2025kan}. A lightweight scale controller predicts dense local scales and normalizes them with an exponential moving average reference. The resulting relative scales transform the queried coordinates, enabling location-specific kernel profiles along a continuous trajectory on a fixed regular sampling grid. An independent mixing controller composes basis responses according to local content, decoupling geometric scale adaptation from content-dependent composition. For efficiency, Core-KAN samples the continuous kernel field at a compact set of scale supports, constructs a response bank, and interpolates neighboring responses according to each local scale. This avoids position-wise kernel synthesis and confines adaptive computation to a compact low-rank basis space with lightweight pointwise projections.

Our contributions are threefold. First, we propose Core-KAN, a relative-scale-conditioned continuous convolution operator that parameterizes a shared coordinate-to-kernel field with a KAN, enabling smooth location-specific adaptation on a fixed sampling grid. Second, we develop an efficient support-and-interpolate scheme with a dual-control mechanism that decouples geometric scale adaptation from content-dependent basis composition. Finally, extensive evaluations on ImageNet-1K, COCO, and ADE20K demonstrate that Core-KAN is an effective plug-and-play operator with consistent performance gains and interpretable adaptive behavior.

\section{Related Work}
\label{sec:related_work}

\subsection{Adaptive Convolutional Operators}

Adaptive ConvNets enlarge the receptive field, select among filters, or generate input-dependent weights. Selective Kernel Networks and large-kernel models provide multiple or wider spatial supports~\cite{li2019selectivekernel,ding2022replknet,liu2023slak,ding2024unireplknet}. Recent efficient backbones further use selective large kernels, multi-scale kernels, anisotropic strip kernels, and partial-channel processing~\cite{li2023lsknet,cai2024pkinet,yuan2026striprcnn,huang2026partialnet}. RefConv reparameterizes filters from pretrained kernels, while SCConv reduces spatial and channel redundancy~\cite{cai2025refconv,li2023scconv}. These designs improve context modeling or efficiency, but their filters remain finite tensors or branches.

Conditional and dynamic convolution instead adapts learned components from the input. CondConv and Dynamic Convolution combine experts~\cite{yang2019condconv,chen2020dynamicconv}; ODConv attends over spatial, channel, and kernel dimensions~\cite{li2022odconv}; KernelWarehouse assembles kernels from shared cells~\cite{li2024kernelwarehouse}; and FDConv modulates frequency-grouped kernel parameters~\cite{chen2025fdconv}. Spatially varying filters are also produced by Dynamic Filter Networks, DDF, pixel-adaptive convolution, and involution~\cite{jia2016dynamicfilter,zhou2021ddf,su2019pac,li2021involution}.

\subsection{Sampling Adaptation and Continuous Kernels}

Deformable convolution approaches spatial adaptivity from the sampling side. By predicting offsets and modulation weights, they allow the operator to collect evidence from irregular positions around each location~\cite{dai2017deformable,zhu2019dcnv2,xiong2024dcnv4}. This mechanism is powerful when object geometry suggests that the sampling grid itself should move. Core-KAN follows a different principle. It keeps the regular convolutional grid intact and adapts the kernel function evaluated on that grid. Thus, the operator preserves the implementation structure and locality bias of convolution while allowing the filter profile to vary continuously.

Continuous kernel methods provide another route to parameter efficiency by representing weights as coordinate-conditioned functions rather than independent lattice parameters. Prior work has shown that coordinate-to-weight mappings can compactly describe kernels for point clouds, long sequences, or large spatial supports~\cite{wang2018pccn,romero2022ckconv,romero2022flexconv,kim2023smpconv}. However, these continuous kernels are usually learned as layer-level or globally controlled functions. They are not primarily designed for dense, location-wise kernel adaptation inside standard 2D visual backbones.

\begin{figure*}[t]
\centering
\includegraphics[width=0.98\textwidth]{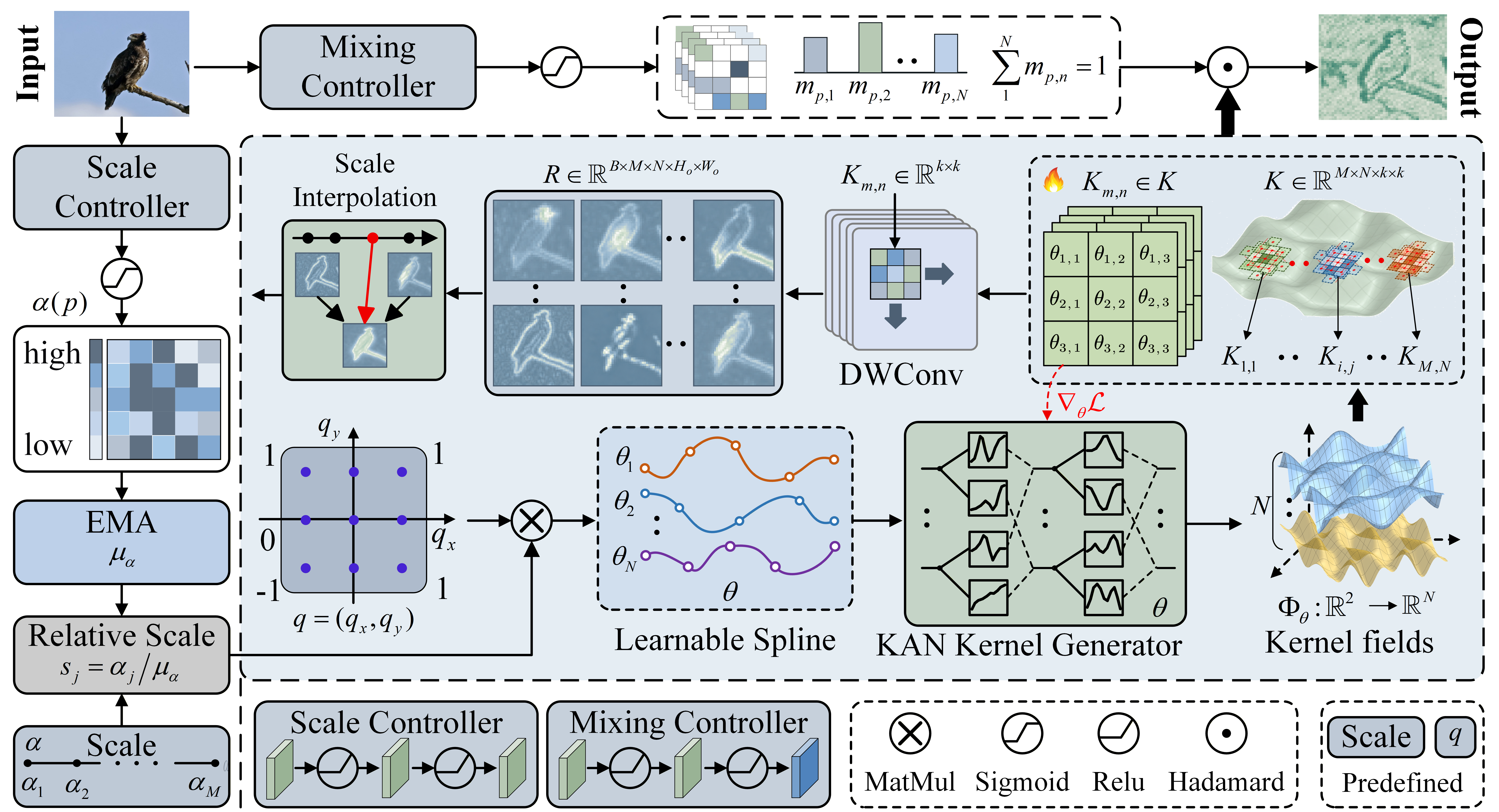}
\caption{Overview of Core-KAN. The scale controller predicts a dense scale field $\alpha(p)$, and an operator-specific EMA reference $\mu_{\alpha}$ converts absolute scale supports into relative supports. These supports transform a predefined coordinate grid $q$ before it is evaluated by the KAN kernel generator. The learnable spline parameters $\theta$ define $N$ shared continuous kernel fields, from which $M$ sets of fixed-size kernels are sampled. Depthwise convolution produces a response bank $\mathcal{R}$, and each location interpolates its two neighboring support responses.}
\label{fig:corekan-overview}
\end{figure*}

% In parallel, the mixing controller predicts simplex weights over the $N$ basis responses. The input and output $1\times1$ projections are omitted from the diagram for clarity.

\subsection{KANs for Visual Modeling}

Kolmogorov-Arnold Networks replace fixed scalar weights and activations with learnable univariate edge functions, giving them a different functional parameterization from conventional multilayer perceptrons~\cite{liu2025kan}. This makes the kernel transformation traceable as combinations of one-dimensional functions. Unlike MLPs, which apply high-dimensional nonlinear changes, each input coordinate's effect on the output can be understood through the shape of its corresponding edge function. This gives the kernel evolution explicit geometric meaning. Existing visual uses of KANs mainly treat them as feature transformation modules, inserting KAN style nonlinear mappings into local or tokenized visual representations~\cite{bodner2024convkan,li2025ukan}. In that setting, the KAN directly transforms image features.

Subsequent studies have broadened this paradigm to other building blocks of visual architectures. For instance, KAT replaces MLP blocks in Transformers with scalable rational-function KAN layers~\cite{yang2025kat}, whereas KAC incorporates a KAN-based classifier for continual visual learning~\cite{hu2025kac}. Despite their emerging potential across vision tasks, leveraging the continuous functional parameterization of KANs to implement efficient, spatially adaptive convolution operators remains largely unexplored.

\section{Method}
\label{sec:method}

We propose \textbf{Core-KAN}, a relative-scale-conditioned continuous convolution operator that decouples geometric scale adaptation from content-dependent basis composition. As illustrated in Figure~\ref{fig:corekan-overview}, a scale controller predicts a dense local scale field, while an exponential moving average (EMA) provides an operator-specific reference for relative-scale normalization. A shared Kolmogorov-Arnold Network (KAN) maps scale-transformed coordinates to continuous kernel weights. Instead of synthesizing a different kernel at every spatial position, Core-KAN samples this continuous field at a compact set of scale supports, constructs a response bank using shared depthwise convolutions, and interpolates neighboring responses according to the local relative scale. An independent mixing controller then performs content-dependent composition over the latent basis responses.

\subsection{Operator Formulation}
\label{sec:corekan-overview}

Let $X\in\mathbb{R}^{B\times C_{\mathrm{in}}\times H\times W}$ and $Y\in\mathbb{R}^{B\times C_{\mathrm{out}}\times H_o\times W_o}$ denote the input and output features. Core-KAN first projects the input into a compact $N$-dimensional basis space:
\begin{equation}
V=P_{\mathrm{in}}(X), \qquad V\in\mathbb{R}^{B\times N\times H\times W},
\label{eq:corekan-input-proj}
\end{equation}
where $P_{\mathrm{in}}$ is a $1\times1$ convolution. The operator computes one scale-adapted response $\widetilde{R}_{b,n}(p)$ for each sample $b$, basis $n$, and output location $p$. It then modulates these responses with content-dependent weights $m_b(p,n)$ and applies a pointwise output projection:
\begin{equation}
Y_{b,c}(p) = \sum_{n=1}^{N} W^{\mathrm{out}}_{c,n}\, m_b(p,n)\, \widetilde{R}_{b,n}(p),
\label{eq:corekan-overall}
\end{equation}
where $W^{\mathrm{out}}$ is the weight matrix of $P_{\mathrm{out}}$. Equation~\eqref{eq:corekan-overall} exposes the two complementary controls in Core-KAN: the scale branch changes the spatial profile of each basis response, whereas the mixing branch changes its content-dependent contribution.

For clarity, the spatial equations below use unit-stride indexing. General stride, padding, and output resolution follow the standard convolutional convention.

\subsection{Decoupled Scale and Mixing Controllers}
\label{sec:corekan-controllers}

Both controllers operate on the original input $X$. In our implementation, each controller contains two lightweight $3\times3$ convolution-GroupNorm-ReLU blocks followed by a $1\times1$ prediction layer.

\paragraph{Dense scale prediction.} The scale controller $C_{\alpha}$ predicts a bounded scalar at every spatial location:
\begin{equation}
\begin{aligned}
\widehat{\alpha}_b(p) &= \sigma\!\left(C_{\alpha}(X_b)(p)\right), \\
\alpha_b(p) &= \alpha_{\min} + \left(\alpha_{\max}-\alpha_{\min}\right) \widehat{\alpha}_b(p),
\end{aligned}
\label{eq:corekan-alpha}
\end{equation}
where $\sigma$ is the sigmoid function and $\alpha_b(p)\in[\alpha_{\min},\alpha_{\max}]$.

Because feature statistics vary across layers and training iterations, the same absolute prediction can have different meanings in different operators. Each Core-KAN layer therefore maintains a non-learnable EMA reference. At training iteration $t$, it is updated by
\begin{equation}
\begin{aligned}
\overline{\alpha}^{(t)} &= \frac{1}{BHW} \sum_{b=1}^{B}\sum_{p} \alpha_b^{(t)}(p), \\
\mu_{\alpha}^{(t)} &= \operatorname{clip}_{[\alpha_{\min},\alpha_{\max}]} \!\left( \rho\mu_{\alpha}^{(t-1)} + (1-\rho) \operatorname{sg}\!\left( \overline{\alpha}^{(t)} \right) \right),
\end{aligned}
\label{eq:corekan-ema}
\end{equation}
where $\rho$ is the EMA momentum and $\operatorname{sg}(\cdot)$ denotes stop-gradient. The stored reference is fixed at inference time. The local relative scale is then
\begin{equation}
s_b(p)=\frac{\alpha_b(p)}{\mu_{\alpha}}.
\label{eq:corekan-relative-scale}
\end{equation}
This normalization expresses the predicted scale relative to the typical scale of the current operator.

\paragraph{Content-dependent mixing.} The independent mixing controller $C_m$ predicts a simplex distribution over the $N$ latent bases:
\begin{equation}
\begin{aligned}
m_b(p,:) &= \operatorname{softmax} \!\left(C_m(X_b)(p)\right), \\
m_b(p,n)&\geq 0, \qquad \sum_{n=1}^{N}m_b(p,n)=1.
\end{aligned}
\label{eq:corekan-mixing}
\end{equation}
The mixing weights depend on local content but do not determine the relative scale used to query the kernel field. This separation prevents the adaptation of the geometric scale and the selection of the bases from being represented by a single routing signal.

\begin{figure}[!t]
\centering
\includegraphics[width=0.9\columnwidth]{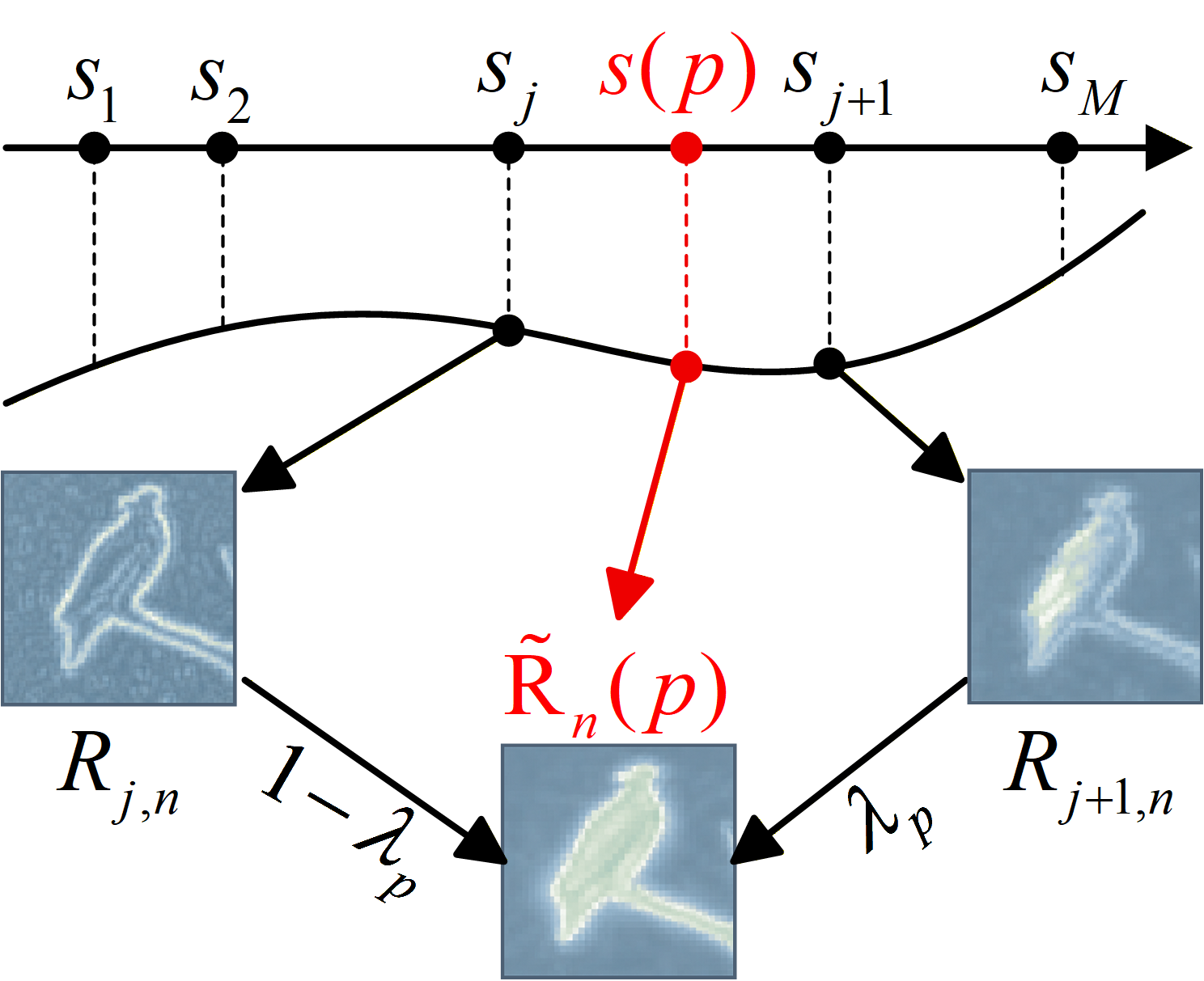}
\caption{Pixel-wise scale interpolation in response space. For a location $p$ with $s_j\leq s(p)\leq s_{j+1}$, Core-KAN retrieves the neighboring support responses $R_{j,n}(p)$ and $R_{j+1,n}(p)$ and linearly combines them using weights $1-\lambda_p$ and $\lambda_p$, respectively. Batch indices are omitted for clarity.}
\label{fig:scale-interpolation}
\end{figure}

\subsection{KAN-Parameterized Continuous Kernel Fields}
\label{sec:corekan-kernel-field}

\paragraph{Relative-scale supports.} We place $M$ uniformly spaced absolute supports in the bounded prediction interval:
\begin{equation}
\alpha_j = \alpha_{\min} + \frac{j-1}{M-1} \left(\alpha_{\max}-\alpha_{\min}\right), \qquad j=1,\ldots,M.
\label{eq:corekan-absolute-supports}
\end{equation}
The supports are converted to relative coordinates using the same EMA reference as the dense scale map:
\begin{equation}
s_j=\frac{\alpha_j}{\mu_{\alpha}}, \qquad j=1,\ldots,M.
\label{eq:corekan-relative-supports}
\end{equation}
Thus, $s_b(p)$ and $\{s_j\}_{j=1}^{M}$ lie in the same operator-calibrated scale coordinate system.

\paragraph{Continuous coordinate-to-kernel mapping.} Let $\mathcal{G}_k=\{\delta_r\}_{r=1}^{k^2}$ denote the offsets of a regular $k\times k$ convolutional grid, and let $q_r=(q_{x,r},q_{y,r})\in[-1,1]^2$ be the normalized coordinate associated with $\delta_r$. A shared KAN maps a continuous two-dimensional coordinate to $N$ scalar kernel fields:
\begin{equation}
\Phi_{\theta}:\mathbb{R}^{2}\rightarrow\mathbb{R}^{N}.
\label{eq:corekan-kan-map}
\end{equation}
For a KAN with $L$ layers, the $\ell$-th layer can be written as
\begin{equation}
z_v^{(\ell+1)} = \sum_{u=1}^{d_{\ell}} \phi_{v,u}^{(\ell)} \!\left(z_u^{(\ell)}\right), \qquad z^{(0)}=x,
\label{eq:corekan-kan-layer}
\end{equation}
where each $\phi_{v,u}^{(\ell)}$ is a learnable univariate edge function parametrized by spline. The symbol $\theta$ collectively denotes all learnable spline parameters, and $\Phi_{\theta}(x)=z^{(L)}$.

For a relative scale $s$, Core-KAN evaluates the same KAN at the scale-transformed coordinates $\{s q_r\}_{r=1}^{k^2}$. The unnormalized $n$-th basis kernel is
\begin{equation}
G_n(s) = \operatorname{reshape}_{k\times k} \!\left( \left\{ \left[\Phi_{\theta}(s q_r)\right]_n \right\}_{r=1}^{k^2} \right).
\label{eq:corekan-continuous-kernel-raw}
\end{equation}
Each sampled kernel is standardized over its spatial entries and rescaled to a Kaiming-compatible magnitude:
\begin{equation}
K_n(s) = \sqrt{\frac{2}{k^2}}\, \frac{ G_n(s)-\operatorname{mean}\!\left(G_n(s)\right) }{ \operatorname{std}\!\left(G_n(s)\right)+\varepsilon }.
\label{eq:corekan-kernel-norm}
\end{equation}
Sampling the continuous fields at the $M$ relative supports yields
\begin{equation}
\mathcal{K} = \left\{ K_{j,n}=K_n(s_j) \right\}_{\substack{j=1,\ldots,M\\n=1,\ldots,N}} \in \mathbb{R}^{M\times N\times k\times k}.
\label{eq:corekan-kernel-bank}
\end{equation}
The kernels in $\mathcal{K}$ are therefore samples from the same shared continuous fields, rather than $M N$ independently stored filters. Moreover, the discrete offsets $\delta_r$ remain fixed for all scales. Varying $s$ changes the coordinates queried from the continuous kernel field, and hence the kernel weights, but does not resize or deform the $k\times k$ sampling grid.

\subsection{Support-and-Interpolate Readout}
\label{sec:corekan-readout}

Directly querying $K_n(s_b(p))$ at every spatial location would require synthesizing a distinct kernel for each position. Core-KAN instead evaluates the kernel field only at the $M$ supports and realizes dense adaptation through response-space interpolation.

\paragraph{Support-sampled response bank.} Each support kernel is applied depthwise to the projected basis feature $V$:
\begin{equation}
R_{b,j,n}(p) = \sum_{r=1}^{k^2} K_{j,n}(\delta_r)\, V_{b,n}(p+\delta_r).
\label{eq:corekan-response-bank-element}
\end{equation}
Stacking the responses over all supports and bases gives
\begin{equation}
\mathcal{R} \in \mathbb{R}^{B\times M\times N\times H_o\times W_o}.
\label{eq:corekan-response-bank}
\end{equation}
This response bank requires only $M$ shared depthwise convolutions in the compact basis space.

\paragraph{Pixel-wise scale interpolation.} If the convolution changes spatial resolution, the scale map is first bilinearly resized to $(H_o,W_o)$. We reuse $\alpha_b(p)$ and $s_b(p)$ for the resized values. For the neighboring supports satisfying $\alpha_j\leq\alpha_b(p)\leq\alpha_{j+1}$, equivalently $s_j\leq s_b(p)\leq s_{j+1}$, the interpolation coefficient is
\begin{equation}
\lambda_{b,p} = \frac{\alpha_b(p)-\alpha_j} {\alpha_{j+1}-\alpha_j} = \frac{s_b(p)-s_j} {s_{j+1}-s_j}, \qquad \lambda_{b,p}\in[0,1].
\label{eq:corekan-interp-weight}
\end{equation}
The equality follows because the local prediction and all support values share the same positive reference $\mu_{\alpha}$. The scale-adapted response is then
\begin{equation}
\widetilde{R}_{b,n}(p) = \left(1-\lambda_{b,p}\right)R_{b,j,n}(p) + \lambda_{b,p}R_{b,j+1,n}(p).
\label{eq:corekan-response-interp}
\end{equation}
Values outside the supported interval are clamped to its nearest boundary; at an exact support location, the interpolation reduces to that support response.

To relate the efficient readout to direct continuous evaluation, define
\begin{equation}
R^{\star}_{b,n}(p) = \sum_{r=1}^{k^2} K_n\!\left(s_b(p)\right)(\delta_r)\, V_{b,n}(p+\delta_r),
\label{eq:corekan-direct-readout}
\end{equation}
which evaluates the continuous kernel field separately at every position. Because convolution is linear in the kernel weights, Equation~\eqref{eq:corekan-response-interp} is exactly equivalent to applying the interpolated local kernel
\begin{equation}
\widetilde{K}_{b,n,p} = \left(1-\lambda_{b,p}\right)K_{j,n} + \lambda_{b,p}K_{j+1,n},
\label{eq:corekan-interp-kernel}
\end{equation}
namely,
\begin{equation}
\widetilde{R}_{b,n}(p) = \sum_{r=1}^{k^2} \widetilde{K}_{b,n,p}(\delta_r)\, V_{b,n}(p+\delta_r).
\label{eq:corekan-local-aggregation}
\end{equation}
Thus, response interpolation exactly realizes the piecewise-linearly interpolated kernel in Equation~\eqref{eq:corekan-interp-kernel}, without materializing a location-specific kernel tensor. Relative to the direct query in Equation~\eqref{eq:corekan-direct-readout}, it is a controllable piecewise-linear approximation along the learned scale-conditioned kernel trajectory. Increasing $M$ improves the sampling density, while a compact $M$ reduces response-bank computation.

\begin{table}[!t]
    \centering
    \footnotesize
    \setlength{\tabcolsep}{2pt}

    \begin{tabular*}{\linewidth}
        {@{\extracolsep{\fill}}lccc@{}}
        \toprule
        Method
        & Params (M) $\downarrow$
        & Top-1 (\%) $\uparrow$
        & Top-5 (\%) $\uparrow$ \\
        \midrule

        ResNet-50\venue{CVPR'16}
        & 25.56 & 78.44 & 94.24 \\

        DY-Conv\venue{CVPR'20}
        & 100.88 & 79.00 & 94.27 \\

        ODConv\venue{ICLR'22}
        & 90.67 & 78.52 & 94.01 \\

        % LSKNet\venue{ICCV'23}
        % & 19.20 & 78.96 & 94.36 \\

        % PKINet\venue{CVPR'24}
        % & 18.70 & 79.12 & 94.51 \\

        % StripNet\venue{AAAI'26}
        % & 16.30 & 80.14 & 95.00 \\

        SCConv\venue{CVPR'23}
        & 17.69 & 79.89 & 94.76 \\

        RefConv\venue{TNNLS'25}
        & 36.97 & 79.91 & 94.61 \\

        PartialNet\venue{AAAI'26}
        & 18.00 & 80.61 & 95.13 \\

        KernelWarehouse\venue{ICML'24}
        & 102.02 & \underline{81.05} & \underline{95.21} \\

        FDConv\venue{CVPR'25}
        & 29.20 & 80.36 & 95.02 \\

        \textbf{Core-KAN (Ours)}
        & 26.61 & \textbf{81.45} & \textbf{95.68} \\

        \bottomrule
    \end{tabular*}

    \caption{
        ImageNet-1K validation results. Reported or reproduced training recipes differ across methods, so the table is intended as a reference comparison. \textbf{Bold} and \underline{underlined} values indicate the best and second-best accuracy, respectively.
    }
    \label{tab:imagenet_main}
\end{table}

\begin{table*}[t]
    \centering
    \footnotesize
    \setlength{\tabcolsep}{2.0pt}
    \renewcommand{\arraystretch}{1.08}

    \begin{tabular*}{\textwidth}
        {@{\extracolsep{\fill}}lcccccccccccc@{}}
        \toprule
        \multirow{2}{*}{Method}
        & \multicolumn{6}{c}{Object Detection}
        & \multicolumn{6}{c}{Instance Segmentation} \\
        \cmidrule(lr){2-7}
        \cmidrule(lr){8-13}

        & $\mathrm{AP}^{\mathrm{box}}$
        & $\mathrm{AP}^{\mathrm{box}}_{50}$
        & $\mathrm{AP}^{\mathrm{box}}_{75}$
        & $\mathrm{AP}^{\mathrm{box}}_{\mathrm{S}}$
        & $\mathrm{AP}^{\mathrm{box}}_{\mathrm{M}}$
        & $\mathrm{AP}^{\mathrm{box}}_{\mathrm{L}}$
        & $\mathrm{AP}^{\mathrm{mask}}$
        & $\mathrm{AP}^{\mathrm{mask}}_{50}$
        & $\mathrm{AP}^{\mathrm{mask}}_{75}$
        & $\mathrm{AP}^{\mathrm{mask}}_{\mathrm{S}}$
        & $\mathrm{AP}^{\mathrm{mask}}_{\mathrm{M}}$
        & $\mathrm{AP}^{\mathrm{mask}}_{\mathrm{L}}$ \\
        \midrule

        ResNet-50\venue{CVPR'16}
        & 37.9 & 58.7 & 41.2 & 21.6 & 41.5 & 49.3
        & 34.5 & 55.6 & 36.8 & 15.9 & 37.1 & 50.4 \\

        DY-Conv\venue{CVPR'20}
        & 39.2 & 60.3 & 42.5 & 23.0 & 42.9 & 51.4
        & 34.7 & 56.0 & 37.1 & 16.4 & 36.9 & 51.1 \\

        ODConv\venue{ICLR'22}
        & 40.1 & 61.5 & 43.6 & 24.0 & 43.6 & 52.3
        & 36.7 & 58.5 & 39.6 & 18.6 & 39.0 & 52.8 \\

        KernelWarehouse\venue{ICML'24}
        & 42.4 & \textbf{65.4} & \underline{46.3}
        & \textbf{27.2} & 46.2 & 54.6
        & \underline{38.9} & \underline{62.0} & \underline{41.5}
        & \textbf{22.7} & \underline{42.6} & 53.1 \\

        FDConv\venue{CVPR'25}
        & \underline{42.5} & 64.8 & 46.2
        & 26.4 & \underline{47.0} & \underline{54.9}
        & 38.3 & 61.8 & 41.0
        & 19.6 & 42.4 & \underline{54.3} \\

        \textbf{Core-KAN (Ours)}
        & \textbf{43.0} & \underline{65.1} & \textbf{47.0}
        & \underline{27.1} & \textbf{47.1} & \textbf{55.8}
        & \textbf{39.5} & \textbf{62.1} & \textbf{42.4}
        & \underline{20.5} & \textbf{42.8} & \textbf{57.2} \\

        \bottomrule
    \end{tabular*}
    \caption{COCO val2017 results using Mask R-CNN under the standard $1\times$ schedule (12 epochs). \textbf{Bold} and \underline{underlined} values denote the best and second-best results, respectively; ties are marked equally.}
    \label{tab:coco_1x}
\end{table*}

\begin{table*}[t]
    \centering
    \footnotesize
    \setlength{\tabcolsep}{2.0pt}
    \renewcommand{\arraystretch}{1.08}

    \begin{tabular*}{\textwidth}
        {@{\extracolsep{\fill}}lcccccccccccc@{}}
        \toprule
        \multirow{2}{*}{Method}
        & \multicolumn{6}{c}{Object Detection}
        & \multicolumn{6}{c}{Instance Segmentation} \\
        \cmidrule(lr){2-7}
        \cmidrule(lr){8-13}

        & $\mathrm{AP}^{\mathrm{box}}$
        & $\mathrm{AP}^{\mathrm{box}}_{50}$
        & $\mathrm{AP}^{\mathrm{box}}_{75}$
        & $\mathrm{AP}^{\mathrm{box}}_{\mathrm{S}}$
        & $\mathrm{AP}^{\mathrm{box}}_{\mathrm{M}}$
        & $\mathrm{AP}^{\mathrm{box}}_{\mathrm{L}}$
        & $\mathrm{AP}^{\mathrm{mask}}$
        & $\mathrm{AP}^{\mathrm{mask}}_{50}$
        & $\mathrm{AP}^{\mathrm{mask}}_{75}$
        & $\mathrm{AP}^{\mathrm{mask}}_{\mathrm{S}}$
        & $\mathrm{AP}^{\mathrm{mask}}_{\mathrm{M}}$
        & $\mathrm{AP}^{\mathrm{mask}}_{\mathrm{L}}$ \\
        \midrule

        ResNet-50\venue{CVPR'16}
        & 40.9 & 61.3 & 44.8 & 24.4 & 44.6 & 52.3
        & 37.1 & 58.3 & 39.9 & 18.4 & 39.8 & 52.9 \\

        DY-Conv\venue{CVPR'20}
        & 41.9 & 63.0 & 45.7 & 25.8 & 45.6 & 53.8
        & 36.9 & 58.5 & 39.7 & 18.8 & 39.4 & 53.2 \\

        ODConv\venue{ICLR'22}
        & 42.6 & 64.0 & 46.6 & 27.0 & 46.3 & 54.8
        & 38.7 & 60.6 & 42.0 & 21.4 & 41.2 & 54.7 \\

        KernelWarehouse\venue{ICML'24}
        & 45.6 & \underline{67.5} & 49.8
        & 29.8 & 49.4 & \textbf{59.0}
        & 41.5 & \underline{63.0} & 44.8
        & \underline{24.2} & 43.9 & 58.5 \\

        FDConv\venue{CVPR'25}
        & \underline{45.7} & 67.3 & \underline{50.4}
        & \underline{30.5} & \underline{49.7} & 58.4
        & \underline{41.7} & 62.5 & \underline{45.0}
        & 24.0 & \underline{44.2} & \underline{58.7} \\

        \textbf{Core-KAN (Ours)}
        & \textbf{46.2} & \textbf{67.8} & \textbf{50.5}
        & \textbf{31.3} & \textbf{49.8} & \underline{58.9}
        & \textbf{42.2} & \textbf{63.8} & \textbf{45.1}
        & \textbf{24.8} & \textbf{44.5} & \textbf{59.2} \\

        \bottomrule
    \end{tabular*}
    \caption{COCO val2017 results using Mask R-CNN under the standard $3\times$ schedule (36 epochs). \textbf{Bold} and \underline{underlined} values denote the best and second-best results, respectively; ties are marked equally.}
    \label{tab:coco_3x}
\end{table*}

\begin{table}[!t]
    \centering
    \footnotesize
    \setlength{\tabcolsep}{1.0pt}
    \renewcommand{\arraystretch}{1.05}

    \begin{tabular*}{\columnwidth}
        {@{\extracolsep{\fill}}lccc@{}}
        \toprule
        Method
        & Params (M) $\downarrow$
        & mIoU (\%) $\uparrow$
        & mAcc (\%) $\uparrow$ \\
        \midrule

        ResNet-50\venue{CVPR'16}
        & 66.00 & 40.00 & 49.61 \\

        DY-Conv\venue{CVPR'20}
        & 140.00 & 41.20 & 51.05 \\

        ODConv\venue{ICLR'22}
        & 131.00 & 42.36 & 52.51 \\

        KernelWarehouse\venue{ICML'24}
        & 141.00 & 43.20 & 53.30 \\

        FDConv\venue{CVPR'25}
        & 70.00 & \underline{43.50} & \underline{53.63} \\

        \textbf{Core-KAN (Ours)}
        & 68.50 & \textbf{44.19} & \textbf{54.38} \\

        \bottomrule
    \end{tabular*}
    \caption{ADE20K validation results with a ResNet-50 backbone. Params denotes the complete segmentation model. \textbf{Bold} and \underline{underlined} values denote the best and second-best results, respectively.}
    \label{tab:ade20k_main}
\end{table}

\subsection{Content Mixing and Low-Rank Interpretation}
\label{sec:corekan-mixing}

The interpolated basis responses encode geometric scale adaptation. Core-KAN subsequently applies the independent content weights from Equation~\eqref{eq:corekan-mixing}:
\begin{equation}
Z_{b,n}(p) = m_b(p,n)\, \widetilde{R}_{b,n}(p).
\label{eq:corekan-gating}
\end{equation}
When required, the mixing map is bilinearly resized to $(H_o,W_o)$ and renormalized over $n$. The output is $Y=P_{\mathrm{out}}(Z)$, yielding Equation~\eqref{eq:corekan-overall}.

The complete operator admits a low-rank interpretation. Let $W^{\mathrm{in}}$ and $W^{\mathrm{out}}$ denote the input and output projection matrices. Combining Equations~\eqref{eq:corekan-input-proj}, \eqref{eq:corekan-local-aggregation}, and \eqref{eq:corekan-overall} gives the effective location-dependent kernel between input channel $c'$ and output channel $c$:
\begin{equation}
W^{\mathrm{eff}}_{b,c,c',p}(\delta) = \sum_{n=1}^{N} W^{\mathrm{out}}_{c,n}\, m_b(p,n)\, \widetilde{K}_{b,n,p}(\delta)\, W^{\mathrm{in}}_{n,c'}.
\label{eq:corekan-effective-kernel}
\end{equation}
Here, $s_b(p)$ controls the spatial profile $\widetilde{K}_{b,n,p}$, whereas $m_b(p,n)$ controls the content-dependent composition of the latent bases. The surrounding pointwise projections share channel-mixing factors across locations, confining the spatially adaptive computation to the compact $N$-dimensional basis space.

% \begin{figure*}[t]
%     \centering
%     \includegraphics[width=\textwidth]{image/ade20k_alpha_mixing_maps_two_best.png}
%     \caption{Scale and mixing behavior on two high-gain ADE20K examples. Columns show the input, ground truth, ResNet-50 and Core-KAN predictions, log-relative scale, dominant basis, normalized mixing entropy, and error difference. Green pixels are corrected by Core-KAN and orange pixels are regressions. Values below each row give the per-image mIoU change.}
%     \label{fig:alpha_mixing_maps}
% \end{figure*}

\begin{figure*}[!t]
    \centering
    \includegraphics[width=\textwidth]{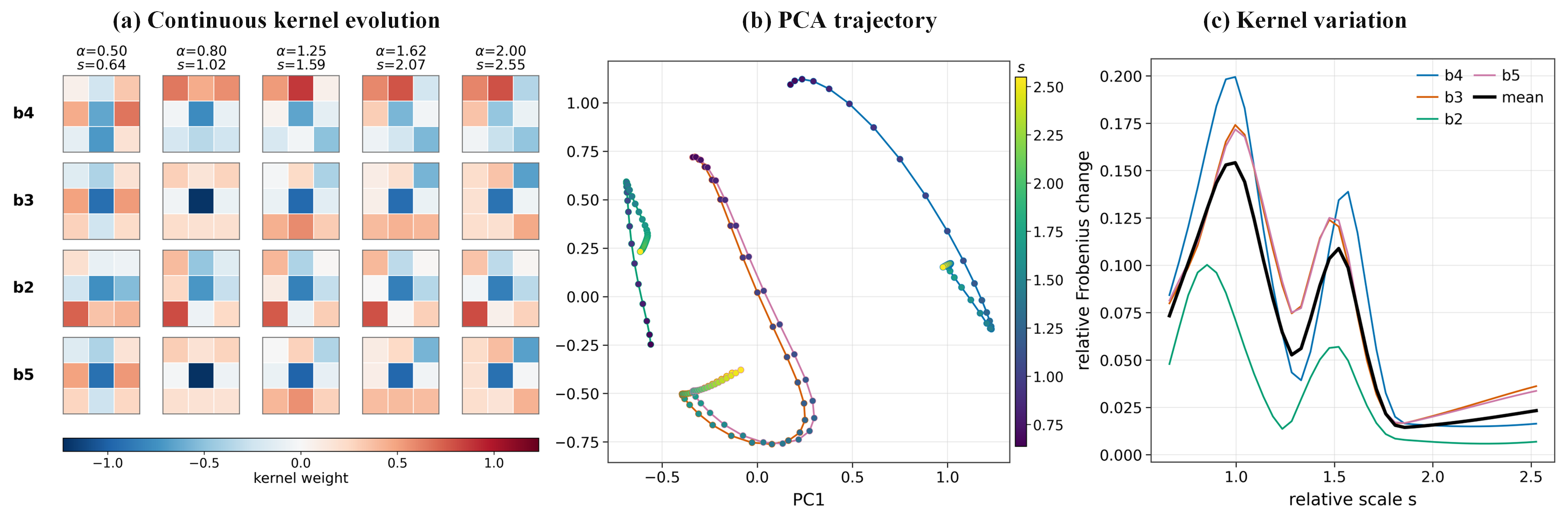}
    \caption{Evolution of learned $3\times3$ kernels in the last Core-KAN block. (a) Four basis fields queried at five relative scales. (b) PCA trajectories of densely sampled kernels. (c) Relative Frobenius change between adjacent queries. Smooth, basis-specific trajectories indicate a continuous kernel field rather than a discrete lookup table.}
    \label{fig:kernel_evolution}
\end{figure*}

\begin{figure}[!t]
    \centering
    \includegraphics[width=\columnwidth]{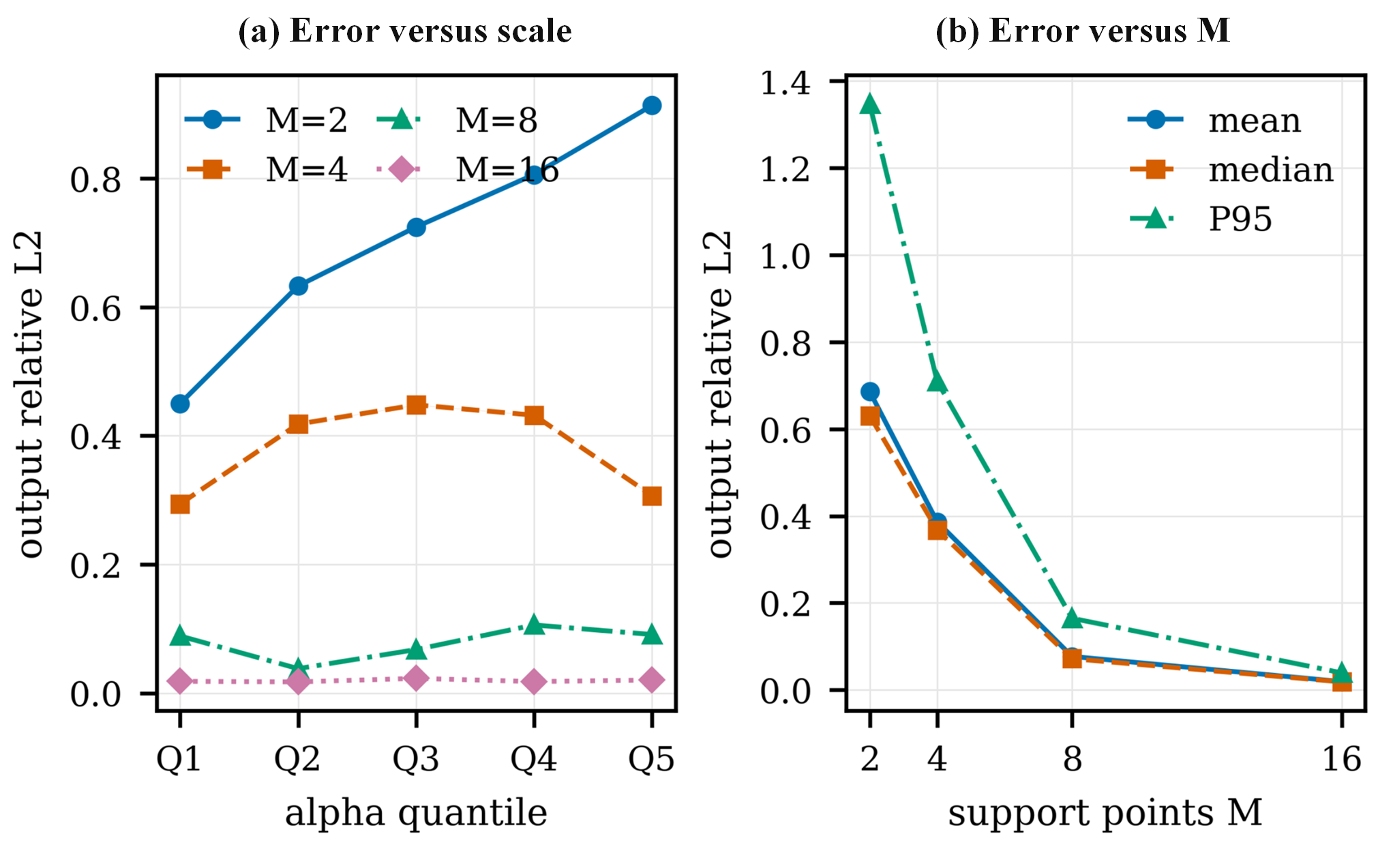}
    \caption{Interpolation fidelity to direct continuous-kernel evaluation. (a) Relative output error across five scale quantiles for different support counts $M$. (b) Mean, median, and 95th-percentile error. Denser supports consistently reduce the approximation error.}
    \label{fig:interpolation_fidelity}
\end{figure}

\section{Experiments}
\label{sec:experiments}

\subsection{Experimental Setup}
\label{sec:experimental_setup}

\paragraph{Core-KAN configuration.} Unless otherwise specified, Core-KAN uses a fixed spatial support of $k=3$, $N=16$ continuous kernel basis fields, and $M=8$ uniformly spaced scale supports. The predicted local scale is bounded to $[\alpha_{\min},\alpha_{\max}]=[0.5,2.0]$, and the momentum of the exponential moving average scale reference is set to $\rho=0.99$. Both the scale controller and the content-mixing controller use a hidden width of 32. The continuous kernel generator consists of two KAN layer with a grid size of 7 and a spline order of 3.  For details, see Appendix A.2-A.6 and E.1-E.4.

\paragraph{Training and evaluation.} Core-KAN is trained on ImageNet-1K~\cite{russakovsky2015imagenet} for 300 epochs with AdamW, a batch size of 4,096, an initial learning rate of $4\times10^{-3}$, 20 warm-up epochs, cosine decay, and single-crop $224\times224$ validation. COCO 2017~\cite{lin2014microsoftcoco} uses Mask R-CNN~\cite{he2017maskrcnn} with FPN~\cite{lin2017fpn} under the standard $1\times$ (12-epoch) and $3\times$ (36-epoch) schedules. For ADE20K~\cite{zhou2017ade20k}, the models are trained for 160K iterations. Within each downstream benchmark, all reproduced backbones share the same data pipeline, schedule, task head, and evaluation protocol.

\subsection{Main Results}
\label{sec:main_results}

Table~\ref{tab:imagenet_main} compares Core-KAN with dynamic convolution operators and recent efficient backbones on ImageNet-1K. Core-KAN achieves 81.45\% top-1 and 95.68\% top-5 accuracy with 26.61M parameters. Compared with ResNet-50, it delivers relative improvements of 3.84\% and 1.53\% in top-1 and top-5 accuracy, respectively, with a parameter overhead of only 4.11\%. Core-KAN also provides relative gains of 0.49\% in both accuracy metrics over KernelWarehouse while using 73.92\% fewer parameters.

On COCO, Core-KAN consistently improves Mask R-CNN under both training schedules. As shown in Table~\ref{tab:coco_1x}, Core-KAN achieves 43.0 $\mathrm{AP}^{\mathrm{box}}$ and 39.5 $\mathrm{AP}^{\mathrm{mask}}$, corresponding to relative improvements of 13.46\% and 14.49\% over ResNet-50, respectively. Compared with the strongest competing methods, Core-KAN further improves box AP over FDConv by 1.18\% and mask AP over KernelWarehouse by 1.54\%. The improvements are consistent across object scales: relative to ResNet-50, Core-KAN achieves gains of 25.46\%, 13.49\%, and 13.18\% in $\mathrm{AP}^{\mathrm{box}}_{\mathrm{S}}$, $\mathrm{AP}^{\mathrm{box}}_{\mathrm{M}}$, and $\mathrm{AP}^{\mathrm{box}}_{\mathrm{L}}$, respectively, together with corresponding gains of 28.93\%, 15.36\%, and 13.49\% for mask prediction. Under the longer $3\times$ schedule, Core-KAN obtains 46.2 $\mathrm{AP}^{\mathrm{box}}$ and 42.2 $\mathrm{AP}^{\mathrm{mask}}$ (Table~\ref{tab:coco_3x}), yielding relative improvements of 12.96\% and 13.75\% over ResNet-50, respectively. It also outperforms FDConv, the strongest competing method in terms of overall AP under this schedule, by 1.09\% in box AP and 1.20\% in mask AP. Core-KAN ranks first on five of the six box metrics and all six mask metrics. The only exception is $\mathrm{AP}^{\mathrm{box}}_{\mathrm{L}}$, where its result is only 0.17\% lower than that of KernelWarehouse (58.9 versus 59.0). These results demonstrate that the advantages of Core-KAN remain consistent under longer training and across detection and instance-segmentation tasks.

% On COCO, Core-KAN achieves 43.0 $\mathrm{AP}^{\mathrm{box}}$ and 39.5 $\mathrm{AP}^{\mathrm{mask}}$ under the $1\times$ schedule (Table~\ref{tab:coco_1x}), corresponding to relative improvements of 13.46\% and 14.49\% over ResNet-50. It also surpasses the strongest competing results by 1.18\% in box AP and 1.54\% in mask AP. The improvements remain consistent across small, medium, and large instances, reaching 25.46\%/13.49\%/13.18\% for box AP and 28.93\%/15.36\%/13.49\% for mask AP relative to ResNet-50. Under the $3\times$ schedule, Core-KAN obtains 46.2 $\mathrm{AP}^{\mathrm{box}}$ and 42.2 $\mathrm{AP}^{\mathrm{mask}}$ (Table~\ref{tab:coco_3x}), yielding relative gains of 12.96\% and 13.75\% over ResNet-50 and 1.09\% and 1.20\% over the strongest alternatives, respectively. Core-KAN ranks first on five of the six box metrics and all six mask metrics; the only exception is large-object box AP, where it is marginally lower than KernelWarehouse by 0.17%.

On ADE20K, Core-KAN achieves 44.19\% mIoU and 54.38\% mAcc with 68.50M parameters (Table~\ref{tab:ade20k_main}). These results represent relative improvements of 10.47\% and 9.61\% over ResNet-50, respectively, with a parameter overhead of only 3.79\% in the complete segmentation model. Core-KAN further improves mIoU and mAcc over FDConv by 1.59\% and 1.40\%, respectively, while using 2.14\% fewer parameters.

\subsection{Continuous Kernels and Interpolation}
\label{sec:Continuous_Kernels_and_Interpolation}
% \subsection{Analyses and Discussion}
% \label{sec:Analyses_and_Discussion}

Figure~\ref{fig:kernel_evolution} examines whether the KAN represents a continuous kernel family and, critically, whether this representation provides interpretable kernel evolution. Queries at increasing relative scales change both the signs and spatial arrangements of the normalized weights while retaining a fixed 3 × 3 sampling grid. The ordered, basis-specific PCA trajectories reveal that each basis function captures a distinct and interpretable pattern of scale-dependent kernel transformations. Dense queries form ordered, basis-specific PCA trajectories, demonstrating that the KAN learns semantically meaningful continuous trajectories rather than arbitrary discrete lookup tables.

Figure~\ref{fig:interpolation_fidelity} compares support-based interpolation with direct per-location continuous-kernel evaluation. The approximation error decreases consistently as the number of supports increases across all scale quantiles and summary statistics. The default setting of $M=8$ substantially reduces the error relative to smaller support sets, while $M=16$ provides further improvement at additional response-bank cost. Thus, eight supports offer a practical trade-off between approximation fidelity and computational efficiency.  For details, see Appendix B.1-B.4.

% Together, these analyses validate two central design assumptions of Core-KAN: the KAN parameterizes continuous nonlinear kernel trajectories, and a compact response bank can reliably approximate direct continuous-kernel evaluation. Here, continuous scale denotes a continuously varying kernel-weight profile on a fixed sampling grid, rather than a variable-size kernel lattice.

% Figure~\ref{fig:kernel_evolution} visualizes the learned kernel fields at different relative scales. Although the $3\times3$ sampling grid remains fixed, each basis exhibits smooth, scale-dependent changes in its weight profile. The ordered PCA trajectories further indicate basis-specific continuous evolution rather than independent discrete kernels.

% Figure~\ref{fig:interpolation_fidelity} evaluates the support-based approximation to direct continuous-kernel evaluation. Increasing $M$ consistently reduces the error across scale quantiles and summary statistics. The default $M=8$ provides a favorable balance between approximation fidelity and response-bank cost, while $M=16$ offers further accuracy at higher computation. Together, these results support both the continuity of the learned kernel fields and the effectiveness of response-bank interpolation.

\section{Conclusion}

Core-KAN introduces dense spatial adaptation through a shared continuous kernel field. Relative-scale queries change kernel geometry, a separate controller mixes latent bases from image content, and support-based interpolation avoids explicit kernel synthesis at every position. Across ImageNet-1K, both COCO schedules, and ADE20K, Core-KAN improves on ResNet-50 with modest parameter overhead. Kernel trajectories and interpolation errors support the intended continuous formulation, while the controller maps show complementary spatial behavior. A remaining limitation is that the response-bank cost grows with the number of scale supports; reducing this cost without weakening interpolation fidelity is a useful direction for further work.

\bibliography{ref}

% Check whether the conference requires a reproducibility checklist to be included in the paper.
% If so, you can uncomment the following line and ajust the path to include it.
% \input{ReproducibilityChecklist.tex}

\end{document}